\documentclass[letterpaper]{article} % DO NOT CHANGE THIS
\usepackage{aaai2027}  % DO NOT CHANGE THIS
\nocopyright
\usepackage[hyphens]{url}  % DO NOT CHANGE THIS
\usepackage{graphicx} % DO NOT CHANGE THIS
\usepackage{natbib}  % DO NOT CHANGE THIS AND DO NOT ADD ANY OPTIONS TO IT
\usepackage{caption} % DO NOT CHANGE THIS AND DO NOT ADD ANY OPTIONS TO IT
\usepackage{algorithm}
\usepackage{algorithmic}
\usepackage{amsmath}
\usepackage{amssymb}

\usepackage{newfloat}
\usepackage{listings}
\DeclareCaptionStyle{ruled}{labelfont=normalfont,labelsep=colon,strut=off} % DO NOT CHANGE THIS
\floatstyle{ruled}
\newfloat{listing}{tb}{lst}{}
\floatname{listing}{Listing}

\usepackage{booktabs}

\title{ASUMOT: Motion-Consistency-Based Asynchronous UAV Detection and Tracking with Event Cameras}
\author {
    Baofeng Jia\textsuperscript{\rm 1}\textsuperscript{\rm 2},
    Xiaoyu Chen\textsuperscript{\rm 1}\textsuperscript{\rm 2}\corresponding,
    Jingyuan Zhang\textsuperscript{\rm 1}\textsuperscript{\rm 2},
    Zongze Wu\textsuperscript{\rm 1}\textsuperscript{\rm 2},
    Haochen li\textsuperscript{\rm 1}\textsuperscript{\rm 2},
    Jing Han\textsuperscript{\rm 1}\textsuperscript{\rm 2},
    Lianfa Bai\textsuperscript{\rm 1}\textsuperscript{\rm 2}
}
\affiliations {
    \textsuperscript{\rm 1}State key Laboratory of Extreme Environment Optoelectronic Dynamic Testing Technology and Instrument, Nanjing University of Science and Technology, Nanjing 210094, China\\
    \textsuperscript{\rm 2}Jiangsu Key Laboratory of Visual Sensing and Intelligent Perception, Nanjing University of Science and Technology, Nanjing 210094, China\\
    baofengjia@njust.edu.cn
}

\begin{document}

\maketitle

\begin{abstract}
Event cameras offer microsecond-level temporal resolution and high dynamic range for low-altitude UAV perception. However, long-range UAVs often produce sparse, fragmented, and noise-contaminated event responses, where one semantic target may appear as multiple spatially separated blobs. Direct blob-level asynchronous tracking therefore suffers from duplicate trajectories and unstable identities. We propose ASUMOT, a motion-consistency-based asynchronous UAV detection and tracking framework operating directly on raw events. ASUMOT models each UAV as a set of motion-consistent event blobs. A local motion-consistency estimator triggers reliable candidates, a lightweight multi-task verifier provides UAV confidence and motion-direction cues, and motion-consistency clustering aggregates fragmented blobs into identity-consistent UAV tracks. We also introduce ES-UAV, a high-definition event-level UAV benchmark with dense semantic annotations. Experiments on public UAV tracking data and ES-UAV show that ASUMOT improves the accuracy--efficiency trade-off while preserving asynchronous event processing. Code and Dataset will be released.

\end{abstract}

\section{Introduction}
The rapid proliferation of unmanned aerial vehicles (UAVs) has intensified the demand for reliable low-altitude surveillance. In anti-UAV scenarios, long-range targets are often small, fast, and visually ambiguous, making real-time detection--tracking under cluttered backgrounds challenging. Frame-based methods~\cite{li2021dense,gong2021effective,huang2023anti,sun2024multi} suffer from motion blur, illumination changes, and large inter-frame displacement in dynamic environments.

Event cameras provide microsecond temporal resolution and high dynamic range~\cite{gallego2020event}. Synchronous event methods convert events into frames, voxel grids, or dense representations~\cite{peng2024scene,chen2025event,gehrig2023recurrent,wang2025object}, but this introduces latency--accuracy trade-offs and discards fine-grained timing. Fully asynchronous trackers~\cite{apps2025asynchronous} preserve event-level timing, yet typically promote each salient blob as an independent object. This blob-level assumption is restrictive for UAVs: one semantic UAV may generate multiple fragmented, intermittent, and spatially separated event blobs, causing duplicate trajectories, missed targets, and unstable identities.
\begin{figure}[t]
  \centering
  \includegraphics[width=0.99\columnwidth]{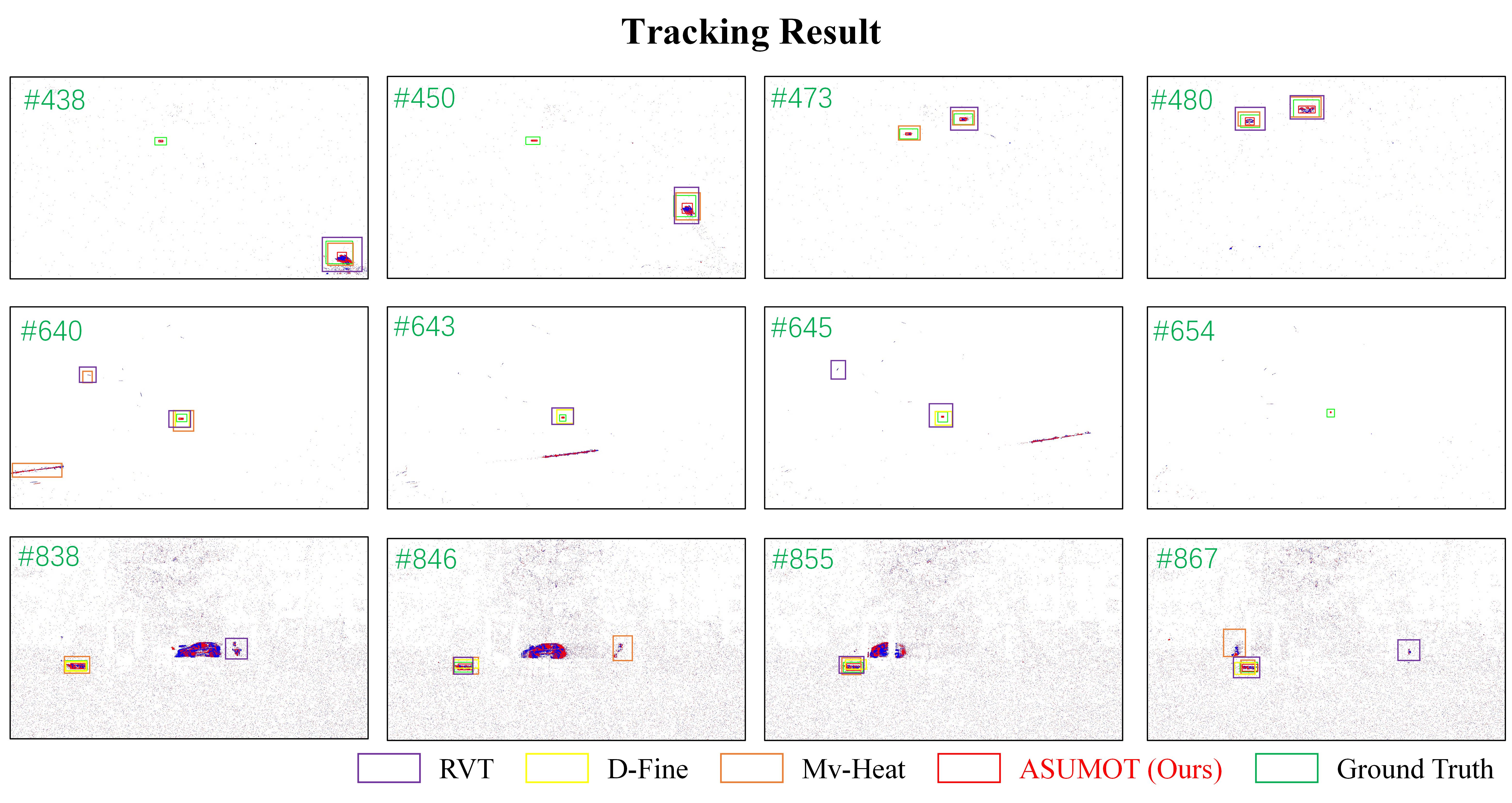}
  \includegraphics[width=0.49\columnwidth]{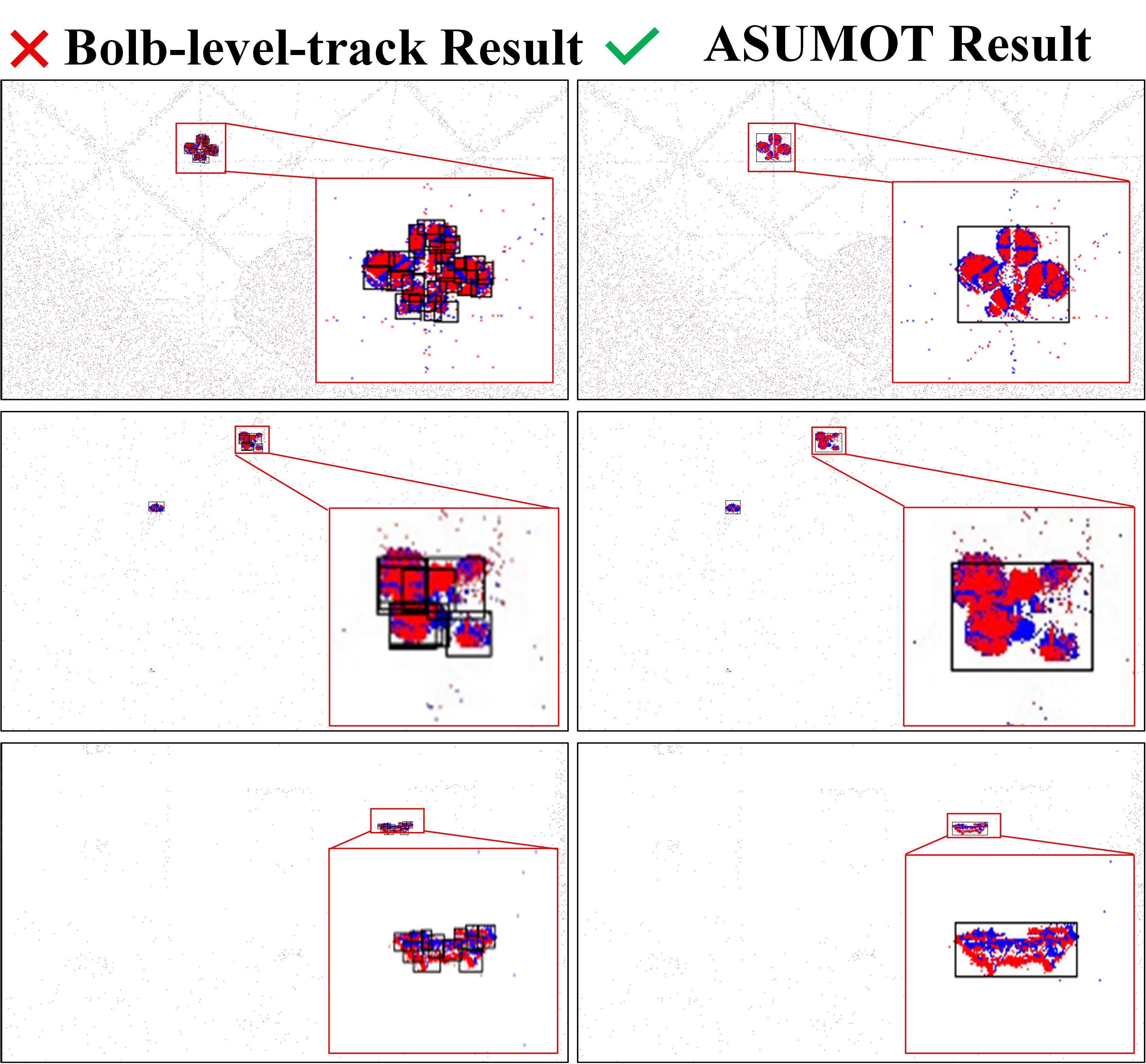}\hfill
  \includegraphics[width=0.49\columnwidth]{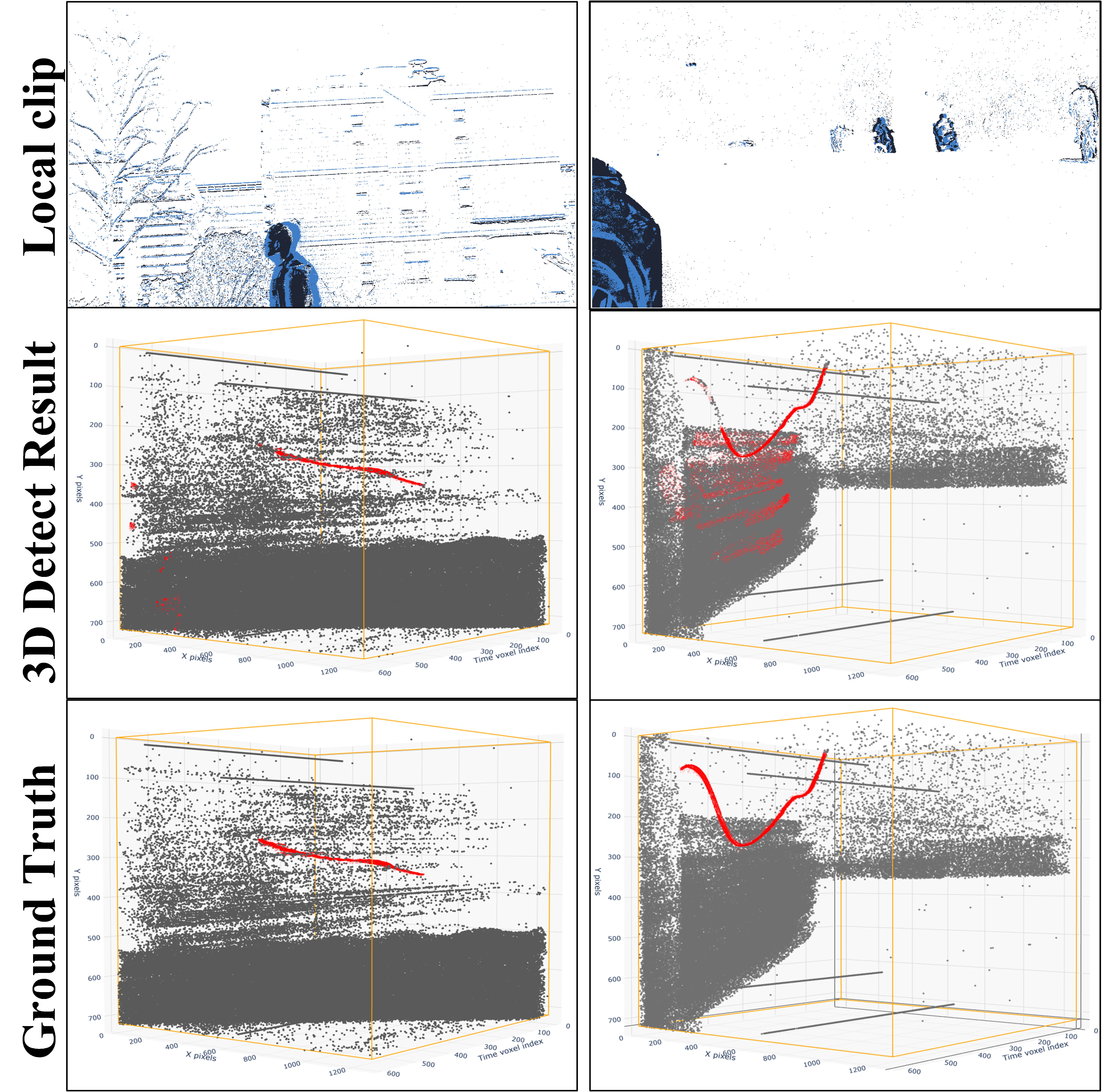}
  \caption{Long-range UAVs produce fragmented blobs, causing duplicate/unstable blob-level tracks. ASUMOT aggregates motion-consistent blobs for stable identity tracking.}
  \label{fig:tracking_results}
\end{figure}

we propose ASUMOT, an asynchronous UAV detection and tracking framework operating directly on raw events. The central idea is to separate low-level event-blob hypotheses from high-level UAV identities. ASUMOT does not assume that one event blob corresponds to one physical target. Instead, each UAV is represented as a time-varying set of motion-consistent event blobs. Low-level blob hypotheses are generated and updated asynchronously, while high-level UAV identities are maintained by aggregating fragmented but motion-consistent blob observations.

We also construct ES-UAV, a high-definition event-level UAV benchmark. Its blob-to-event annotation strategy reduces background mislabeling while preserving target contours and motion trajectories, enabling fine-grained event-domain evaluation. Experiments on public UAV tracking data and ES-UAV show that ASUMOT achieves a favorable accuracy--efficiency trade-off and practical edge-device latency. Our contributions are:
\begin{itemize}
\item We propose ASUMOT, a raw-event asynchronous framework that models each UAV as a set of motion-consistent event blobs instead of a single blob-level track.
\item We design a motion-consistency-driven association and clustering mechanism to aggregate fragmented blob observations into stable UAV-level tracks, reducing duplicate trajectories and identity fragmentation.
\item We construct ES-UAV, a high-definition event-level UAV benchmark with blob-to-event annotations, and validate ASUMOT through comprehensive experiments and edge deployment.
\end{itemize}

% \begin{table*}[t]
%   \centering
%   \small
%   \begin{tabular}{@{}lllll@{}}
%     \toprule
%     Dataset & Label Type & Resolution & Temporal Granularity & Multi-Scene  \\
%     \midrule
%     F-UAV-D~\cite{mandula2024towards} 
%       & BBox & 1280$\times$720 (HD) & Frame-level & $\times$  \\
%     NeRDD~\cite{magrini2024neuromorphic} 
%       & BBox & 1280$\times$720 (HD) & Frame-level & $\times$  \\
%     FRED~\cite{magrini2025fred} 
%       & BBox & 1280$\times$720 (HD) & Frame-level & $\checkmark$   \\
%     EV-UAV~\cite{chen2025event} 
%       & Seg. & 346$\times$260 & Ms-level sparse & $\checkmark$   \\
%     Ours 
%       & Seg. & 1280$\times$720 (HD) & Event-level & $\checkmark$   \\
%     \bottomrule
%   \end{tabular}
%   \caption{Comparison with existing event-based UAV detection datasets. ES-UAV provides high-definition event-level semantic annotations for UAV detection evaluation.}
%   \label{tab:event_uav_datasets}
% \end{table*}
\begin{table*}[t]
  \centering
  \small
  \begin{tabular}{@{}lllccc@{}}
    \toprule
    Dataset & Label Type & Resolution & Temporal Granularity & Multi-Scene & Event Annotation Format\\
    \midrule
    Event-VOT~\cite{wang2024event}& BBox& 1280$\times$720 (HD) & Frame-level &$\checkmark$&$\times$\\
    F-UAV-D~\cite{mandula2024towards} 
      & BBox & 1280$\times$720 (HD) & Frame-level & $\times$ &$\times$ \\
    NeRDD~\cite{magrini2024neuromorphic} 
      & BBox & 1280$\times$720 (HD) & Frame-level & $\checkmark$&$\times$ \\
    FRED~\cite{magrini2025fred} 
      & BBox & 1280$\times$720 (HD) & Frame-level & $\checkmark$&$\times$   \\
    EV-UAV~\cite{chen2025event} 
      & Seg. & 346$\times$260 & Ms-level sparse & $\checkmark$ &Frame-projected  \\
    M$^{2}$E-UAV~\cite{yan2026m} & Seg. & 1280$\times$720 (HD) & Event-level & $\checkmark$&Frame-projected\\
    Ours & Seg. & 1280$\times$720 (HD) & Event-level & $\checkmark$ &Blob-projected  \\
    \bottomrule
  \end{tabular}
  \caption{Comparison with existing event-based UAV detection datasets. ES-UAV provides high-definition event-level semantic annotations for UAV detection evaluation.}
  \label{tab:event_uav_datasets}
\end{table*}

\section{Related Work}

\paragraph{Representation-based event detection and tracking.}
A common approach to event-based perception is to convert asynchronous events into dense representations, such as event frames, voxel grids, or time surfaces, and then apply conventional detection or tracking architectures~\cite{chen2018pseudo,iacono2018towards,jiang2019mixed,hu2020learning,messikommer2020event}. Recent methods further improve event representation with adaptive accumulation~\cite{li2022asynchronous,wang2024eas}, recurrent networks~\cite{gehrig2023recurrent}, Transformers or state-space models~\cite{peng2024scene,yang2025smamba,xu2025hybrid}, graph neural networks~\cite{mitrokhin2020learning,schaefer2022aegnn}, and spiking neural networks~\cite{cordone2022object,paredes2019unsupervised,shrestha2018slayer,neftci2019surrogate,aitsam2025event}. These methods can exploit mature image-based architectures, but their performance depends heavily on temporal aggregation. For long-range UAVs with sparse and intermittent events, inappropriate accumulation may blur target structure, mix target events with background activity, or introduce additional latency.

\paragraph{Asynchronous event tracking.}
Asynchronous trackers update target states directly as events arrive, avoiding dense event accumulation and preserving the native temporal resolution of event cameras~\cite{chen2019asynchronous,messikommer2023data,apps2025asynchronous}. AEB-style models describe event blobs with probabilistic spatial distributions and update their states using incoming events, while AEMOT~\cite{apps2025asynchronous} detects and validates salient blobs in a fully asynchronous manner. These methods are efficient and suitable for event-level processing. However, they typically treat each validated blob as an independent object. This one-blob-one-object assumption becomes problematic for long-range UAVs, where one physical target may produce multiple fragmented event blobs. ASUMOT differs from these methods by decoupling blob-level event tracking from UAV-level identity maintenance, allowing multiple motion-consistent blobs to jointly represent one semantic UAV.

\paragraph{Event-based UAV perception and datasets.}
Event-based UAV perception has attracted increasing attention because event cameras are naturally suitable for high-speed and high-dynamic-range scenarios~\cite{magrini2025drone}. Existing studies exploit UAV-specific cues such as propeller motion, high-frequency event signatures, and lightweight event-domain detectors for embedded deployment~\cite{sanket2021evpropnet,zhang2025evdetmav,murray2025propeller,eldeborg2024drone,mandula2024towards,chen2025event,magrini2025ev}. Meanwhile, several event-based UAV datasets have been proposed~\cite{mandula2024towards,magrini2024neuromorphic,magrini2025fred,chen2025event}. However, many of them provide frame-level bounding boxes, sparse labels, low-resolution recordings, or short clips. High-definition event-level annotations for long-sequence UAV detection and target-level identity evaluation remain limited. ES-UAV is designed to fill this gap by providing dense event-level semantic labels and long continuous sequences for asynchronous UAV detection and tracking evaluation.
\section{ES-UAV: Event-Segmentation UAV Dataset}

We introduce ES-UAV, an event-based UAV evaluation dataset. Data are captured using an EVK5 event camera with a spatial resolution of (1280$\times$720) and microsecond-level temporal resolution. ES-UAV is acquired in the form of long sequences, collected in cluttered urban environments with pedestrians, vehicles, buildings, and vegetation. The UAV motions include hovering, abrupt turns, field-of-view entry and exit, and composite maneuvers. In total, ES-UAV contains over 6B events, among which 36M events are densely annotated at the event level.

\begin{figure}[t]
\centering
\includegraphics[width=0.49\columnwidth]{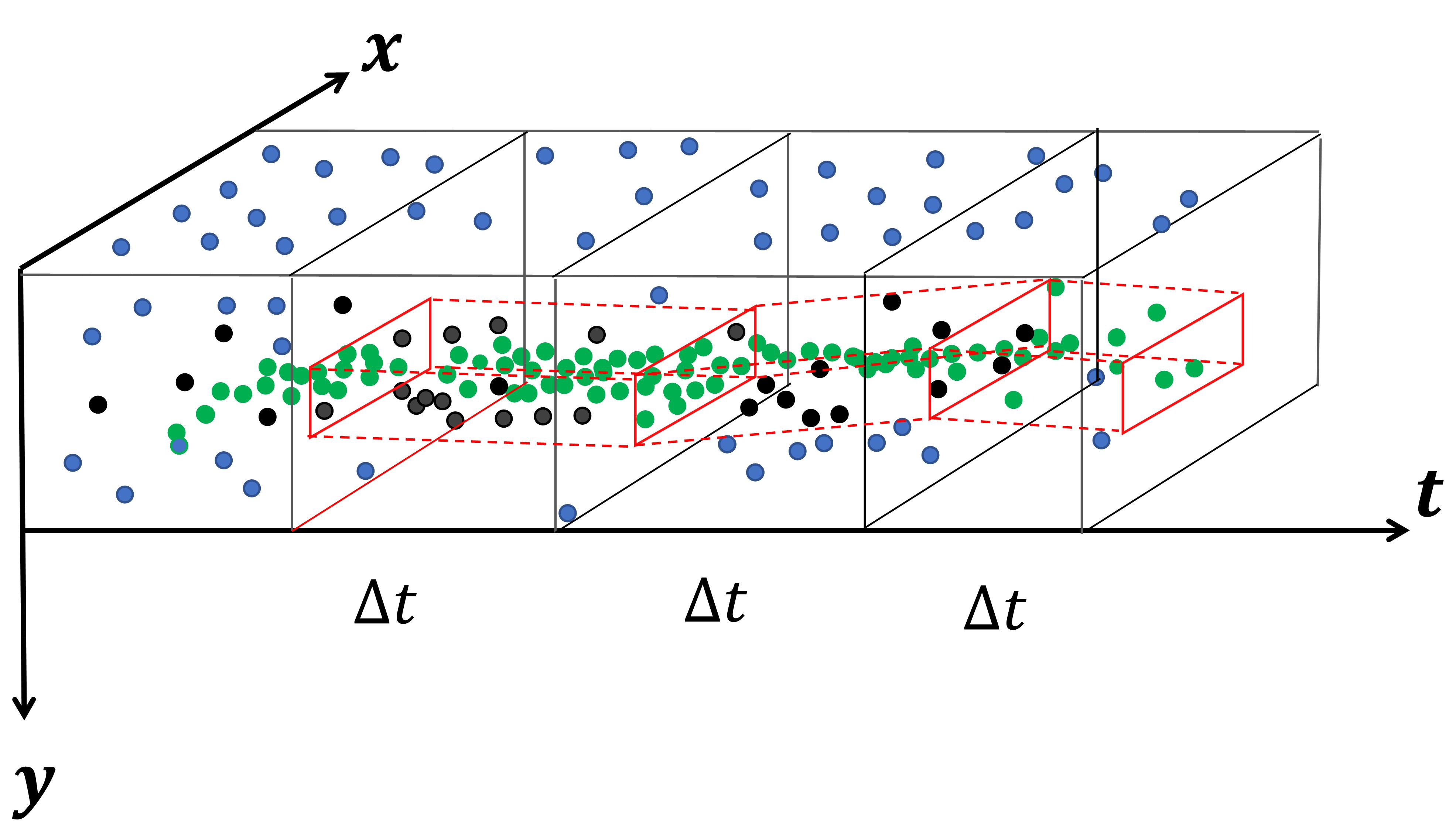}\hfill
\includegraphics[width=0.49\columnwidth]{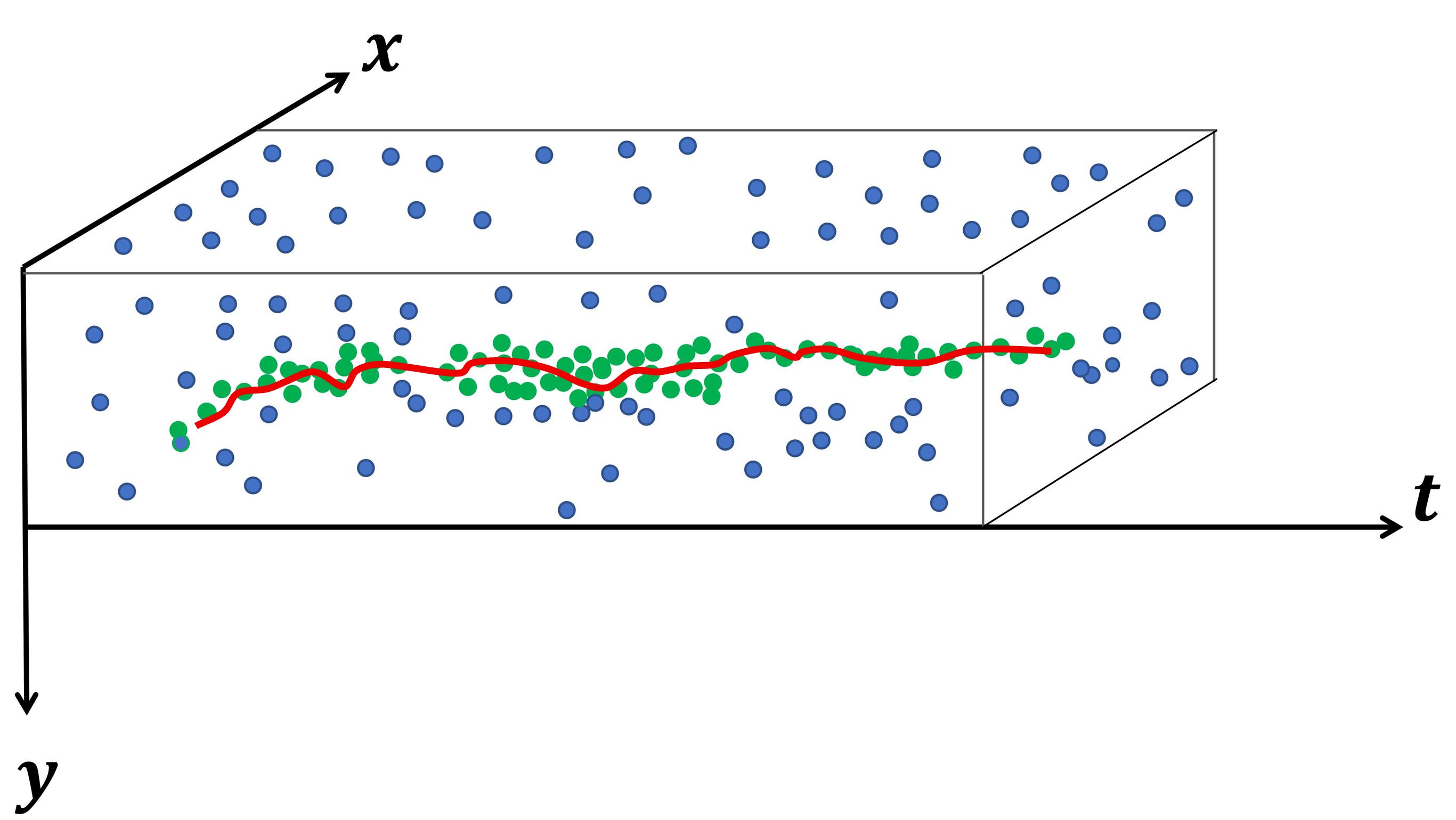}
\makebox[0.49\columnwidth][c]{(a) Frame-to-event}\hfill
\makebox[0.49\columnwidth][c]{(b) Blob-to-event}
\caption{Comparison of annotation strategies. Frame-to-event annotation projects coarse frame-level labels back to events and may introduce background noise. In contrast, blob-to-event annotation initializes asynchronous blob trackers from manually selected seed points, uses multiple blobs to cover fragmented UAV responses, and assigns semantic labels to the corresponding raw events.}
\label{fig:annotation_strategies}
\end{figure}

\paragraph{Blob-to-event annotation.}
Annotating UAVs in raw event streams is difficult because event responses are sparse, asynchronous, and easily mixed with background activity. Conventional frame-level annotation first accumulates events into image-like frames and then labels targets on the accumulated representation. Although this strategy is efficient, it may blur small UAV structures and introduce background events when frame-level labels are projected back to the raw stream. To obtain fine-grained event-level labels, ES-UAV adopts a semi-automatic blob-to-event annotation strategy. Annotators first manually select seed points on UAV event responses, and an asynchronous blob tracker is then initialized from these seeds to follow the corresponding event blobs directly in the raw stream. For complex UAV targets with fragmented or spatially separated responses, multiple blob trackers are used jointly to cover different target parts. During annotation, annotators can add or delete tracking points to correct drift, occlusion, target deformation, and intermittent event responses. The tracked blobs are finally projected back to the original event stream, assigning semantic labels to the raw events covered by valid UAV blobs, as shown in Fig.~\ref{fig:annotation_strategies}.

Formally, given a set of annotated UAV blob trajectories $\{\mathcal{B}_r\}_{r=1}^{R}$, each raw event $e_k=(\boldsymbol{\xi}_k,\sigma_k,t_k)$ is assigned a binary semantic label:
\begin{equation}
y_k =
\begin{cases}
1, & e_k \in \bigcup_{r=1}^{R}\mathcal{B}_r,\\
0, & \text{otherwise}.
\end{cases}
\end{equation}
Here, $\mathcal{B}_r$ denotes the spatiotemporal support of the $r$-th annotated UAV blob trajectory. Multiple blob trajectories may correspond to the same UAV target when the target response is fragmented. This semi-automatic strategy preserves fine UAV contours and motion trajectories while significantly reducing the manual effort compared with event-by-event labeling.

\begin{figure}[t]
\centering
\includegraphics[width=0.99\columnwidth]{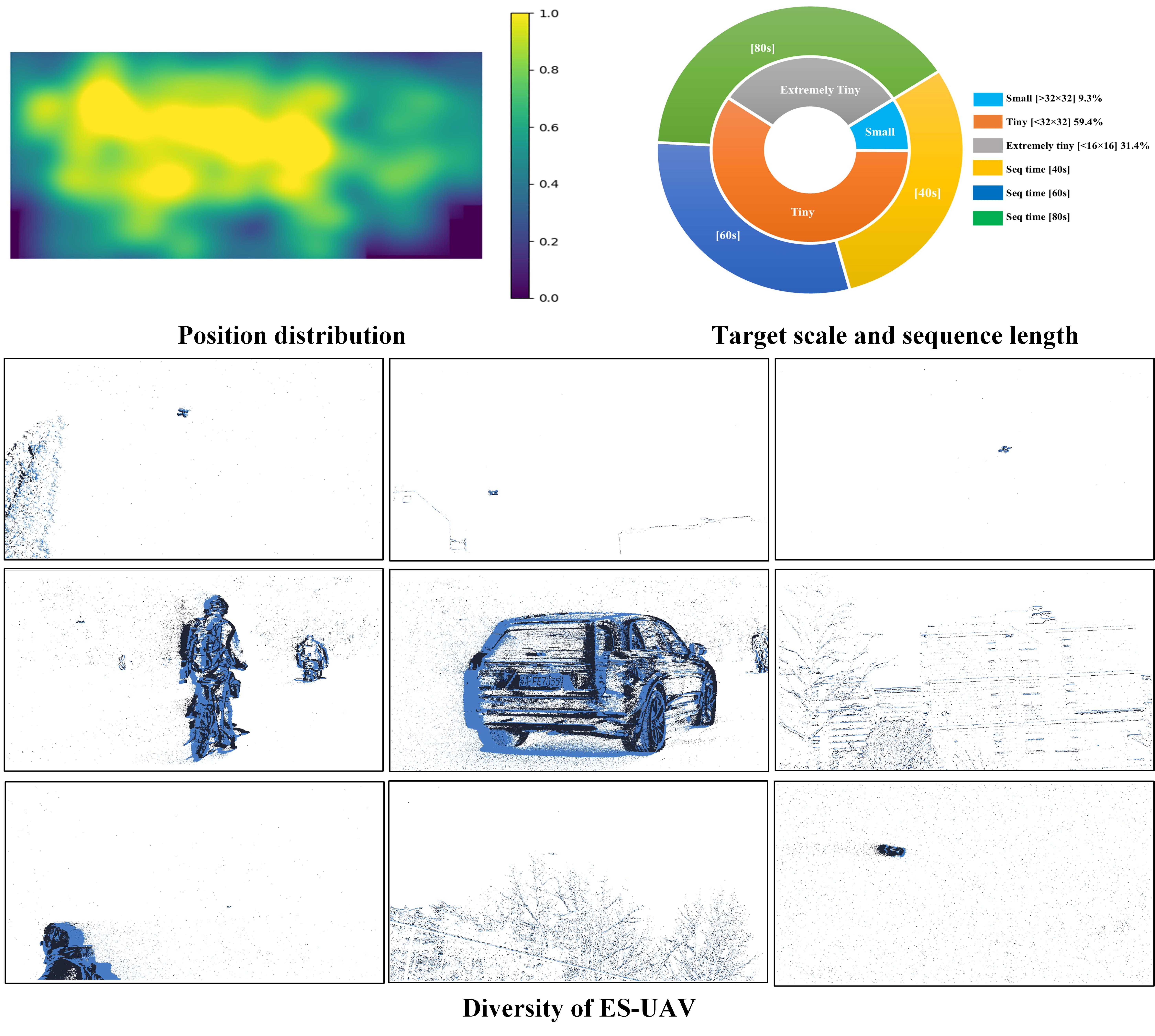}

\caption{Statistics and examples of ES-UAV.}
\label{fig:es_uav_dataset}
\end{figure}

% ES-UAV is designed as a test-only benchmark: evaluated methods are trained on public datasets to assess cross-dataset generalization. As shown in Fig.~\ref{fig:es_uav_dataset}, UAVs appear in diverse complex environments, and most targets are smaller than (32$\times$32) pixels, reflecting the typical characteristics of long‑range and small targets in Anti‑UAV perception. The long continuous sequences further support evaluation of temporal robustness, identity consistency, and high-definition asynchronous UAV detection.
\paragraph{Evaluation protocol.}
ES-UAV is designed as a test-only benchmark. No ES-UAV labels are used for training the verifier, selecting hyperparameters, or tuning thresholds. Learning-based components and learning-based baselines are trained on public event datasets under the same protocol, while ES-UAV is reserved for evaluating cross-dataset generalization. As shown in Fig.~\ref{fig:es_uav_dataset}, UAVs appear in diverse complex environments, and most targets are smaller than $32\times32$ pixels, reflecting the typical long-range small-target setting in anti-UAV perception. The long continuous sequences further support evaluation of temporal robustness, identity consistency, duplicate-track suppression, and high-definition asynchronous UAV detection.

\section{Method}

\subsection{Problem Formulation}

We formulate event-based UAV detection and tracking as an online probabilistic state estimation problem over raw asynchronous events. Following AEMOT~\cite{apps2025asynchronous} and the AEB tracker~\cite{wang2024asynchronous}, an event stream is denoted as
\begin{equation}
\mathcal{E}=\{e_k\}_{k=1}^{K}, \qquad
e_k=(\boldsymbol{\xi}_k,\sigma_k,t_k),
\end{equation}
where $\boldsymbol{\xi}_k=(x_k,y_k)^\top\in\mathbb{R}^2$ is the pixel location, $\sigma_k\in\{-1,+1\}$ is the polarity, and $t_k$ is the timestamp.

The state of the $i$-th event blob at time $t$ is represented as
\begin{equation}
\boldsymbol{\zeta}_i(t)=
\left(
\mathbf{p}_i(t),
\mathbf{v}_i(t),
\theta_i(t),
q_i(t),
\boldsymbol{\lambda}_i(t),
\boldsymbol{\Delta}_i(t)
\right),
\end{equation}
where $\mathbf{p}_i=(p_{x,i},p_{y,i})^\top$ denotes the blob center, $\mathbf{v}_i=(v_{x,i},v_{y,i})^\top$ denotes the translational velocity, $\theta_i$ and $q_i$ characterize the orientation and angular velocity, $\boldsymbol{\lambda}_i=(\lambda_{i,1},\lambda_{i,2})$ specifies the principal-axis spatial scales, and $\boldsymbol{\Delta}_i$ denotes the polarity offset between positive and negative event responses.

Following the AEB event likelihood model, the probability of an incoming event $e_k$ conditioned on blob state $\boldsymbol{\zeta}_i(t_k)$ is written as
\begin{equation}
p(e_k|\boldsymbol{\zeta}_i(t_k))
\propto
\exp
\left(
-\frac{1}{2}
\tilde{\boldsymbol{\xi}}_{k,i}^{\top}
\boldsymbol{\Lambda}_i(t_k)^{-2}
\tilde{\boldsymbol{\xi}}_{k,i}
\right),
\end{equation}
where
\begin{equation}
\tilde{\boldsymbol{\xi}}_{k,i}
=
\boldsymbol{\xi}_k
-
\mathbf{p}_i(t_k)
-
\sigma_k\boldsymbol{\Delta}_i(t_k),
\end{equation}
and
\begin{equation}
\boldsymbol{\Lambda}_i(t)
=
\mathbf{R}(\theta_i(t))
\begin{pmatrix}
\lambda_{i,1}(t) & 0\\
0 & \lambda_{i,2}(t)
\end{pmatrix}
\mathbf{R}(\theta_i(t))^\top .
\end{equation}
Here, $\boldsymbol{\Lambda}_i(t)\in\mathbb{R}^{2\times2}$ is the blob shape-scale matrix, and $\boldsymbol{\Lambda}_i(t)^{-2}$ follows the AEB convention for evaluating the normalized spatial distance of an event to the blob distribution. Incoming events are associated with existing blob hypotheses by evaluating this likelihood.

Although this blob-level formulation enables asynchronous event assignment and state update, it is insufficient for long-range UAV perception. In cluttered anti-UAV scenarios, a single UAV may generate multiple fragmented and spatially separated event blobs due to small target scale, propeller-induced responses, viewpoint variation, and background interference. Directly promoting each validated blob as an independent object track may therefore cause duplicate trajectories and unstable identities.

To address this issue, we represent the $m$-th UAV target as a time-varying set of motion-consistent event blobs:
\begin{equation}
\mathcal{B}_m(t)
=
\left\{
\boldsymbol{\zeta}_i(t)
\mid
i\in\mathcal{I}_m(t)
\right\},
\end{equation}
where $\mathcal{I}_m(t)$ denotes the index set of event blobs assigned to the $m$-th UAV. The goal of ASUMOT is to update blob-level hypotheses and UAV-level tracks online as events arrive, while preserving the microsecond temporal resolution of event streams.

\subsection{Overview}

\begin{figure*}[t]
\centering
\includegraphics[width=0.97\textwidth]{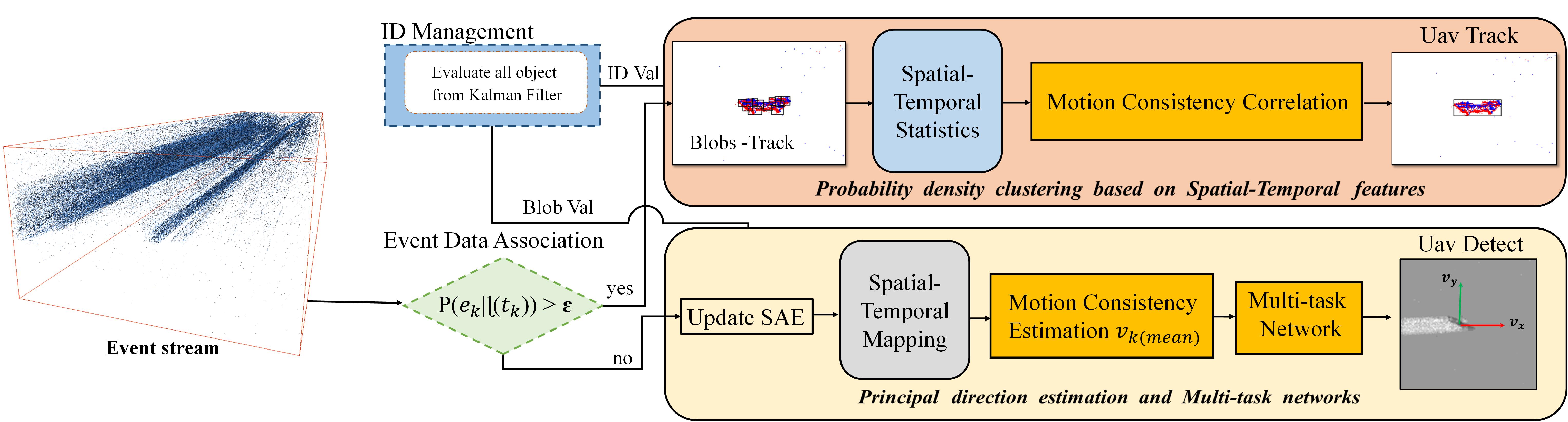}
\caption{Overview of the proposed asynchronous UAV detection and tracking framework. ASUMOT operates directly on raw event streams and consists of event-level association, motion-consistency estimation, candidate verification, dynamic initialization, and UAV-level motion-consistency clustering.}
\label{fig:overview}
\end{figure*}

The overall pipeline of ASUMOT is shown in Fig.~\ref{fig:overview}. For each incoming event, ASUMOT first evaluates whether the event can be associated with existing valid or candidate blob tracks. Unassociated events are used to update the Surface of Active Events (SAE)~\cite{benosman2013event} and trigger candidate generation through motion-consistency estimation. Candidate blobs are further verified by a lightweight multi-task network, which predicts both UAV confidence and motion direction. Verified blobs are then aggregated into UAV-level tracks through Motion-Consistency Clustering (MCC), which associates fragmented but motion-consistent event blobs with the same semantic target. This design preserves asynchronous event processing while reducing duplicate tracking caused by fragmented UAV responses.

The online inference procedure is summarized in Algorithm~\ref{alg:asumot}.
\begin{algorithm}[h]
\caption{Online inference of ASUMOT}
\label{alg:asumot}
\begin{algorithmic}[1]
\REQUIRE Event stream $\mathcal{E}$, blob hypotheses $\mathcal{T}_b$, UAV-level tracks $\mathcal{T}_u$
\FOR{each incoming event $e_k$}
\STATE Predict active blob states to timestamp $t_k$.
\STATE Associate $e_k$ with candidate/valid blobs using event likelihood.
\IF{$e_k$ is uniquely associated}
    \STATE Update the matched blob asynchronously.
\ELSE
    \STATE Update SAE and compute local consistency score $C_k$.
    \IF{$C_k>\tau_c$}
        \STATE Spawn a candidate blob hypothesis.
    \ENDIF
\ENDIF
\STATE Promote reliable candidate blobs using verifier confidence.
\STATE Assign verified blobs to UAV-level tracks using motion-consistency affinity.
\STATE Update UAV identities and remove stale blobs/tracks.
\ENDFOR
\RETURN UAV-level trajectories $\mathcal{T}_u$
\end{algorithmic}
\end{algorithm}

\subsection{Motion-Consistency Modeling for Event Streams}

\subsubsection{Local-Motion-Consistency Estimation.}

\begin{figure}[t]
\centering
\includegraphics[width=0.80\columnwidth]{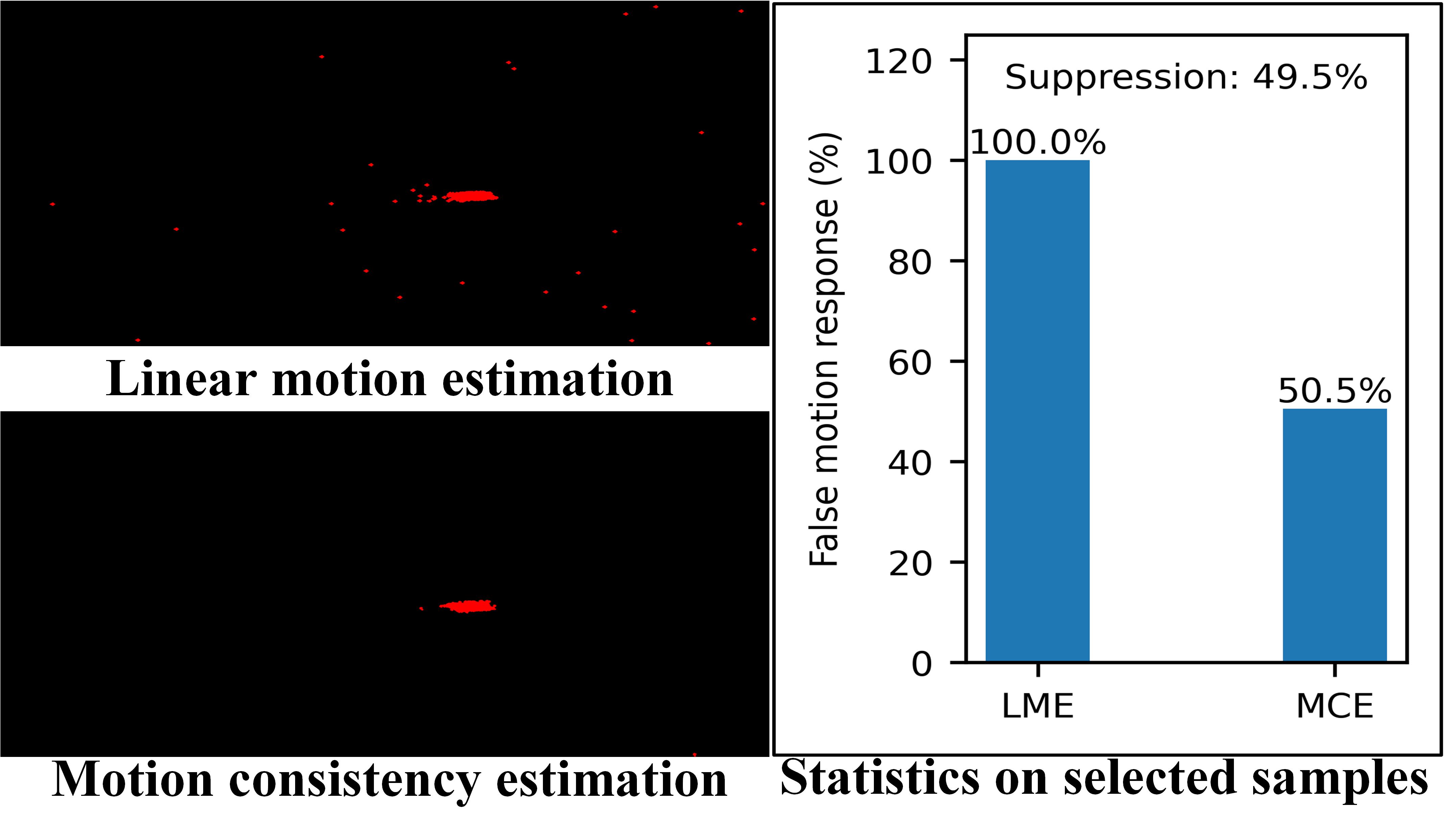}
% \makebox[\columnwidth][c]{(a) Linear motion estimation result}
% \includegraphics[width=0.80\columnwidth]{Figures/87.jpg}
% \makebox[\columnwidth][c]{(b) Motion-consistency estimation result}

\caption{ Comparison between linear motion estimation and motion-consistency estimation. 
Red regions indicate motion responses. Linear motion estimation produces more 
scattered background responses, while motion-consistency estimation suppresses 
noisy responses and preserves more reliable UAV motion cues. Quantitative statistics 
are computed on selected samples, with the baseline normalized to 100\%.}
\label{fig:motion_feature_extraction}
\end{figure}

Within a short temporal interval, UAV-induced events usually form locally coherent structures, while background noise and isolated events are temporally unstable. Instead of estimating dense optical flow, we use the temporal consistency of local event-structure orientations as a lightweight cue for candidate triggering. The extracted orientation is not treated as the physical UAV motion direction, but as a descriptor of the dominant local event structure.

Let $T(\boldsymbol{\xi})$ be the latest timestamp at pixel $\boldsymbol{\xi}$. For an incoming event $e_k$, ORB-oriented ~\cite{rublee2011orb} keypoints are extracted from an SAE patch $\mathcal{P}_k$ centered at $\boldsymbol{\xi}_k$. Each valid keypoint $j\in\mathcal{K}_k$ provides an orientation $\theta_j$ and is assigned a temporal weight:
\begin{equation}
\alpha_j = \exp\left[-\rho\left(t_k-T(\boldsymbol{\xi}_j)\right)\right],
\end{equation}
where $\rho$ is the decay factor. Its structural direction is
\begin{equation}
\mathbf{d}_j=(\cos\theta_j,\sin\theta_j)^\top .
\end{equation}
The reference direction is computed by weighted circular averaging:
\begin{equation}
\mathbf{d}_{\mathrm{ref}} =
\frac{
\sum_{j\in\mathcal{K}_k}\alpha_j\mathbf{d}_j
}{
\left|
\sum_{j\in\mathcal{K}_k}\alpha_j\mathbf{d}_j
\right|_2+\epsilon
}.
\end{equation}
The local consistency score is defined as
\begin{equation}
C_k =
\frac{
\sum_{j\in\mathcal{K}_k}
\alpha_j
\left|
\mathbf{d}_j^\top \mathbf{d}_{\mathrm{ref}}
\right|
}{
\sum_{j\in\mathcal{K}_k}\alpha_j+\epsilon
}.
\end{equation}
A candidate blob is triggered when
\begin{equation}
C_k>\tau_c,
\qquad
|\mathcal{K}_k|>n_{\min},
\end{equation}
where $\tau_c$ is the consistency threshold and $n_{\min}$ is the minimum number of valid keypoints. Fig.~\ref{fig:motion_feature_extraction} show its robustness to background noise. This triggering mechanism suppresses noisy candidates and reduces the burden on subsequent verification and tracking.

\subsubsection{Motion-Consistency Clustering.}

Local motion consistency triggering produces blob-level hypotheses, whereas anti-UAV perception requires target-level identities. Motion-Consistency Clustering (MCC) assigns verified blobs to UAV-level tracks by evaluating whether fragmented blobs are consistent with the same semantic UAV in position, velocity, temporal continuity, and reliability.

For each verified blob $i$ at update step $k$, we define
\begin{equation}
\mathbf{z}_{k,i} =
\left[
\mathbf{p}_{k,i}^{\top},
\mathbf{v}_{k,i}^{\top}
\right]^{\top},
\qquad
\eta_{k,i} = \hat{y}_{k,i} C_{k,i},
\end{equation}
where $\mathbf{p}_{k,i}$ and $\mathbf{v}_{k,i}$ are the blob center and velocity, $\hat{y}_{k,i}$ is the verifier confidence, and $C_{k,i}$ is the local consistency score. The reliability weight $\eta_{k,i}$ suppresses transient clutter and uncertain blobs.

The affinity between blob $i$ and the predicted UAV-level track $m$ is defined as
\begin{equation}
\begin{aligned}
\mathcal{A}_{i,m} = 
&\lambda_p \exp\left(-\frac{\|\mathbf{p}_{k,i}-\mathbf{p}_{k,m}^{-}\|^2}{2\sigma_p^2}\right) \\
&+ \lambda_v \exp\left(-\frac{\|\mathbf{v}_{k,i}-\mathbf{v}_{k,m}^{-}\|^2}{2\sigma_v^2}\right) + \lambda_{\eta}\eta_{k,i} - \lambda_t \Delta t_{i,m}.
\end{aligned}
\end{equation}
where $\mathbf{p}_{k,m}^{-}$ and $\mathbf{v}_{k,m}^{-}$ are the predicted UAV-level position and velocity, and $\Delta t_{i,m}$ is the time since the last update of track $m$. A blob is assigned to the track with the highest affinity if $\mathcal{A}_{i,m}>\tau_a$. Each blob can be assigned to at most one UAV-level track, while each UAV-level track can collect multiple verified blobs.

For the $m$-th UAV-level track, let $\mathcal{I}_{k,m}$ denote its assigned blob set. The aggregated target-level measurement and stability weight are computed as
\begin{equation}
\bar{\mathbf{z}}_{k,m} = \frac{
\sum_{i\in\mathcal{I}_{k,m}}
\eta_{k,i}\mathbf{z}_{k,i}
}{
\sum_{i\in\mathcal{I}_{k,m}}\eta_{k,i}
+\epsilon
},
\qquad
\omega_{k,m} = \frac{
\sum_{i\in\mathcal{I}_{k,m}}\eta_{k,i}
}{
|\mathcal{I}_{k,m}|+\epsilon
}.
\end{equation}
Only stable target-level measurements are used for UAV-level update:
\begin{equation}
\bar{\mathcal{Z}}_k =
\left\{
(\bar{\mathbf{z}}_{k,m},\omega_{k,m})
\mid
\omega_{k,m}>\tau_{\omega}
\right\}.
\end{equation}

The target-level measurements are then passed to a lightweight multi-target filtering backend. In our implementation, we use a weighted GM-PHD~\cite{vo2006gaussian} filter, where $\omega_{k,m}$ modulates the measurement likelihood and identities are maintained by weighted Mahalanobis association. Detailed filtering, pruning, merging, and identity assignment will be provided in the supplementary material. This design keeps the main algorithm centered on blob-to-UAV aggregation while allowing the filtering backend to be replaced by other online multi-object filters.

\subsection{Learning-Based Verification and Dynamic Initialization}

\begin{figure}[t]
\centering
\includegraphics[width=0.92\columnwidth]{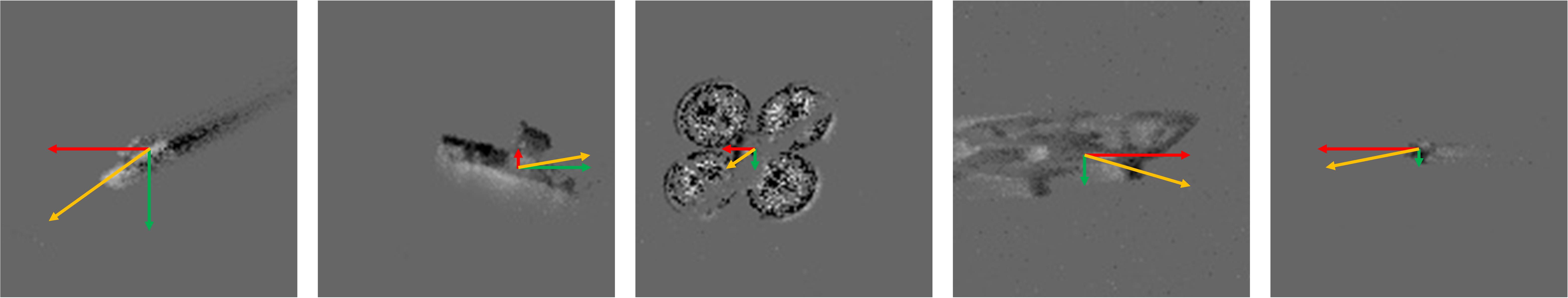}
\makebox[\columnwidth][c]{(a) UAV targets}

\includegraphics[width=0.92\columnwidth]{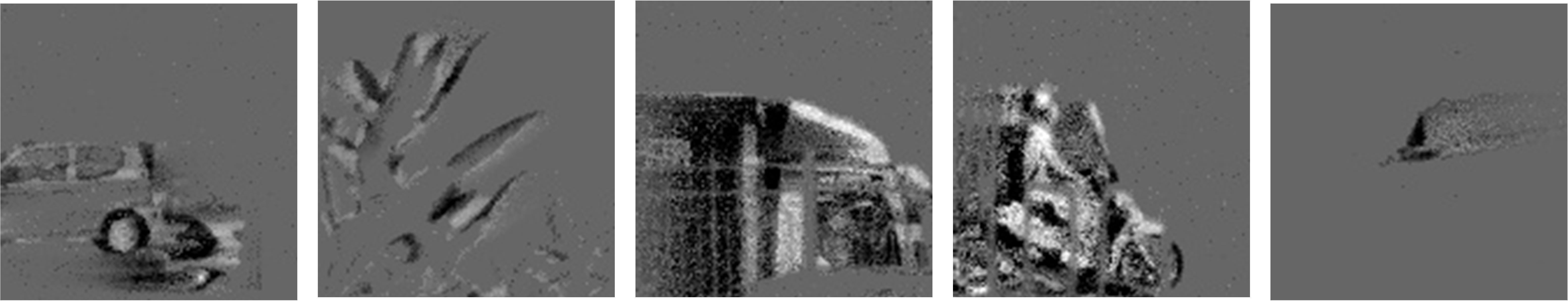}
\makebox[\columnwidth][c]{(b) Non-UAV targets}

\caption{Training samples for the multi-task verifier are collected from the FRED training set. (a) UAV candidates with different viewpoints and scales, where motion directions are visualized. (b) Non-UAV candidates from cluttered urban environments.}
\label{fig:multitask_dataset}
\end{figure}

EKF-based blob trackers are sensitive to initialization under abrupt maneuvers and intermittent events. We therefore use a lightweight multi-task network for candidate verification and dynamic motion initialization. For each candidate blob, an event-intensity patch $I_i\in\mathbb{R}^{128\times128}$ is accumulated in the local track coordinate system with exponential temporal decay. The network $f_{\phi}$ predicts UAV confidence and motion direction:
\begin{equation}
(\hat{y}_i,\hat{\varphi}_i)
=
f_{\phi}(I_i),
\end{equation}
where $\hat{y}_i\in[0,1]$ indicates the probability that the candidate corresponds to a UAV, and $\hat{\varphi}_i$ denotes the estimated motion direction.

It is trained with a joint classification and direction-estimation objective:
\begin{equation}
\mathcal{L}
=
\mathcal{L}_{\mathrm{cls}}
+
\lambda_{\mathrm{dir}}
y_i
\left(
1-\cos(\hat{\varphi}_i-\varphi_i)
\right),
\end{equation}
where $\mathcal{L}_{\mathrm{cls}}$ is the binary cross-entropy loss, $y_i\in\{0,1\}$ is the UAV label, $\varphi_i$ is the ground-truth motion direction, and $\lambda_{\mathrm{dir}}$ balances the two tasks. The direction loss is applied only to positive UAV samples.

During inference, a candidate is promoted when $\hat{y}_i>\tau_y$ over a short buffer. The estimated direction initializes or corrects the blob velocity:
\begin{equation}
\mathbf{v}_i
=
s_i
\left(
\cos\hat{\varphi}_i,
\sin\hat{\varphi}_i
\right)^\top,
\end{equation}
where $s_i$ is the motion magnitude estimated from the local motion-consistency module or inherited from the current tracker state. Candidates with consistently low confidence are deleted, reducing dependence on handcrafted initialization and improving stability.

The verifier uses a compact residual CNN with multi-head outputs and is trained on balanced UAV/non-UAV samples from public event datasets (Fig.~\ref{fig:multitask_dataset}). Since it runs only on candidate patches, its overhead remains low for embedded deployment. Detailed network architecture and FLOPs computation will be provided in the supplementary material.

\subsection{Track Management}

ASUMOT maintains candidate and valid blob tracks. Candidate blobs are generated by motion-consistency triggering and promoted after verification; valid blobs participate in MCC-based UAV-level association. For each event, association is performed first with valid blobs and then with candidates using $p(e_k|\boldsymbol{\zeta}_i)$. A uniquely associated event updates the state asynchronously; a multiply associated event is treated as ambiguous and only prediction is performed; an unmatched event is passed to MCE for candidate generation. Blob tracks are terminated if they leave the image, remain inactive, or are repeatedly rejected. UAV-level tracks are terminated when no motion-consistent blob group is associated for a predefined interval.

\section{Experiment}
\subsection{Datasets and Evaluation Metrics}
We evaluate tracking performance on the FRED dataset and detection performance on the proposed ES-UAV dataset.

\subsubsection{FRED dataset.}
FRED is a multimodal UAV benchmark dataset containing 7 hours of event and RGB recordings across diverse scenes and illumination conditions, supporting both detection and tracking evaluation. Event frames are annotated at 33.33 ms intervals with a resolution of 1280×720. Following the official protocol, we evaluate on the challenging split (147 training clips and 84 testing clips) using event data only. The dataset expands bounding box sizes for modality synchronization; however, since the detection boxes produced by asynchronous methods are computed from event point clouds and thus yield lower IoU with the enlarged labels, we evaluate only the tracking task. Asynchronous outputs are synchronized to the dataset sampling rate to ensure fair comparison.
\subsubsection{ES-UAV dataset.}
ES-UAV is a high-definition event-based UAV evaluation dataset (1280×720, event-level semantic annotations) designed for asynchronous detection accuracy assessment. All methods are trained on public datasets(Frame on FRED and Point on EV-UAV) and tested exclusively on ES-UAV to ensure fair comparison.

\subsubsection{Evaluation Metrics.}
For tracking, we report Multiple Object Tracking Accuracy (MOTA), Identification F1 Score (IDF1), Recall, and Precision under the Euclidean Consistency criterion (EUC = 100) to mitigate synchronization errors caused by fixed-rate evaluation. For detection, we adopt Intersection over Union (IoU), False Alarm rate (Fa), Accuracy (Acc), and Probability of Detection (Pd), following prior event-based UAV segmentation work Ev-SpSegNet~\cite{chen2025event}.

\subsection{Implementation Details}
The system is implemented in C++, with the multi-task network trained in PyTorch and deployed via LibTorch. Training samples are extracted from the FRED training split using 128×128 patches. The network is trained on a single RTX 4090 GPU for 65 epochs with a learning rate of $5\times10^{-4}$. Motion estimation operates on 64×64 local patches. Probability density estimation initializes the filter with multiple hypotheses, and the tracker follows the default configuration in AEB tracker~\cite{wang2024asynchronous}. Edge deployment experiments are conducted on an NVIDIA Jetson TX2.
\subsection{Comparison with State-of-the-Art}
We compare with representative methods across three input representations: event frames, voxel grids, and event streams. For frame- and voxel-based methods, we adopt ByteTrack~\cite{zhang2022bytetrack} as the tracking backend, following the official FRED protocol.
\begin{table*}[t]
  \centering
  \small
  \begin{tabular}{@{}llccccc@{}}
    \toprule
    Method & Data Format & IDF1 (\%) $\uparrow$ & Rcll (\%) $\uparrow$ 
    & Prcn (\%) $\uparrow$ & MOTA (\%) $\uparrow$ & GFLOPs $\downarrow$ \\
    \midrule
    YOLOv11-n~\cite{khanam2024yolov11} 
      & Event-Frame & 13.1 & 51.0 & 95.9 & 48.2 & 6.5 \\
    YOLOv11-s~\cite{khanam2024yolov11} 
      & Event-Frame & 29.0 & 54.8 & 94.8 & 51.6 & 21.5 \\
    MvHeat-DET~\cite{wang2025object} 
      & Event-Frame & 19.5 & 59.0 & 94.8 & 55.1 & 56.4 \\
    D-FINE~\cite{peng2024d} 
      & Event-Frame & 20.0 & 63.1 & 94.9 & 59.2 & 25.0 \\
    SMamba~\cite{yang2025smamba} 
      & Event-Voxel & 37.6 & 62.4 & 94.2 & 56.9 & 7.4 \\
    RVT~\cite{gehrig2023recurrent} 
      & Event-Voxel & \textbf{44.6} & 60.0 & \textbf{96.7} & 56.5 & 8.4 \\
    SAST~\cite{peng2024scene} 
      & Event-Voxel & 38.4 & 61.3 & 95.4 & 56.2 & 6.4 \\
    Ours 
      & Event-Stream & 44.5 & \textbf{79.0} & 84.6 & \textbf{64.0} & \textbf{0.29} \\
    \bottomrule
  \end{tabular}
  \caption{Evaluation results on the challenging split of the FRED dataset under MOTA@EUC=100.}
  \label{tab:fred_results}
\end{table*}

\subsubsection{Results on FRED dataset.}

Table~\ref{tab:fred_results} reports results on the challenging split. ASUMOT achieves the highest MOTA of 64.0\% with only 0.29 GFLOPs. Compared with D-Fine~\cite{peng2024d}, it improves MOTA from 59.2\% to 64.0\% while reducing computation from 25 to 0.29 GFLOPs. The main gain comes from recall: ASUMOT reaches 79.0\%, reducing target loss for small objects, cluttered backgrounds, and rapid motion.Although precision is lower due to more aggressive candidate retention, the substantial recall gain reduces missed targets and improves overall MOTA in fast-moving UAV scenarios.

Fig.~\ref{fig:tracking_results} shows qualitative comparisons. ASUMOT localizes UAVs across varying scales and clutter while maintaining stable identities. Because boxes are inferred from event point clouds rather than accumulated frames, predictions align tightly with small targets. Under temporary disappearance or sparse events, motion-consistent state estimation maintains trajectories without heavy recurrent architectures.

\begin{table}[!h]
  \centering
  \small
  \setlength{\tabcolsep}{4pt}
  \begin{tabular}{@{}lcccccc@{}}
    \toprule
    Method & $P_d$ (\%) $\uparrow$ 
    & $F_a$ ($10^{-4}$) $\downarrow$ 
    & Acc. (\%) $\uparrow$ & IoU (\%) $\uparrow$ \\
    \midrule
   D-FINE &76.0&0.72&57.9&33.5\\
    RVT & 64.0 & 1.02 & 59.0 & 27.5\\
    Ev-SpSegNet & 86.0 & 391.37 & 32.4 & 18.4 \\
    Ours & \textbf{89.9} & \textbf{0.56} & \textbf{93.3} & \textbf{42.8} \\
    \bottomrule
    % \multicolumn{5}{@{}l@{}}{\footnotesize $^\dagger$ Evaluated on 16$\times$ downsampled streams (full-res not supported)}
  \end{tabular}
  \caption{Evaluation results on the ES-UAV dataset.}
  \label{tab:es_uav_results}
\end{table}
\subsubsection{Results on ES-UAV dataset.}
Table~\ref{tab:es_uav_results} summarizes detection performance.Due to the extreme sparsity of UAV event responses, IoU alone cannot fully reflect detection reliability; therefore, Acc and false alarm rate are jointly considered.
ASUMOT achieves the highest Pd (89.9\%), IoU (42.8\%), and Acc (93.3\%) while maintaining the lowest false alarm rate (0.56$\times10^{-4}$). Representation-based approaches such as D-FINE and RVT achieve competitive localization performance by leveraging mature spatial feature extraction, but their temporal aggregation limits precise event-level localization under rapid UAV motion. 
Consequently, they obtain lower Acc and higher false responses compared with ASUMOT. Ev-SpSegNet shows limited generalization on complex backgrounds and high-resolution datasets, highlighting the importance of ES-UAV for evaluating event-based UAV detection tasks.  

Preserving event dynamics and motion consistency is key to reliable UAV perception in clutter. Fig.~\ref{fig:tracking_results} presents qualitative detection results of ASUMOT , and more visual comparisons will be provided in the supplementary material.

\subsubsection{Edge Deployment.}
We build ASUMOT-Tiny with intermittent motion estimation and evaluate it on an edge device.
\begin{table}[t]
  \centering
  \small
  \setlength{\tabcolsep}{5pt}
  \begin{tabular}{@{}lcccccc@{}}
    \toprule
    Method  & MOTA (\%) $\uparrow$ 
    & Latency (ms) $\downarrow$ & GFLOPs $\downarrow$ \\
    \midrule
    YOLOv11-n & 48.2 & \textbf{23.4} & 6.5 \\
    ASUMOT-Tiny  & \textbf{55.0} & 29.94 & \textbf{0.29} \\
    \bottomrule
  \end{tabular}
  \caption{Evaluation results on the challenging split of the FRED dataset on TX2 under MOTA@EUC=100.}
  \label{tab:tx2_results}
\end{table}
Table~\ref{tab:tx2_results} shows that ASUMOT-Tiny achieves 55.0\% MOTA with 29.94 ms median latency on Jetson TX2 at a median input rate of $1.34{\times}10^{5}$ events/s. Compared with synchronous YOLOv11-n, it improves tracking accuracy with comparable runtime and lower computation.

\begin{table}[t]
  \centering
  \small
  \setlength{\tabcolsep}{5pt}
  \begin{tabular}{@{}lccccccc@{}}
    \toprule
    Var. & MCC & MCE & MT & IDF1 & Rcll & Prcn & MOTA \\
    \midrule
    1 & $\times$ & $\checkmark$ & $\checkmark$ 
      & 11.5 & \textbf{85.5} & 19.2 & -274.6 \\
    2 & $\checkmark$ & $\checkmark$ & $\times$ 
      & 34.5 & 70.4 & 76.4 & 47.9 \\
    3 & $\checkmark$ & $\times$ & $\checkmark$ 
      & 36.4 & 73.2 & 79.2 & 53.2 \\
    4 & $\checkmark$ & $\checkmark$ & $\checkmark$ 
      & \textbf{44.5} & 79.0 & \textbf{84.6} & \textbf{64.0} \\
    \bottomrule
  \end{tabular}
  \caption{Ablation study of different components in the proposed ASUMOT model on the FRED dataset. All metrics are reported in percentage. MT denotes the multi-task module(controlling whether dynamic initialization is enabled), MCC denotes motion-consistency clustering, and MCE denotes motion-consistency estimation.}
  \label{tab:ablation_results}
\end{table}

\subsection{Ablation Study}
Table~\ref{tab:ablation_results} validates MCC, MCE, and the multi-task module (MT). The full model achieves the best MOTA/IDF1. Removing MCC causes severe duplicate trajectories and even negative MOTA, which highlights the essential distinction between blob-level trackers and UAV-level target trackers; replacing MCE with linear regression reduces MOTA to 53.2\%, showing the benefit of orientation-consistency filtering; removing MT decreases MOTA to 47.9\%, confirming the value of learning-based verification and motion-direction initialization.
\section{Conclusion}

We have proposed ASUMOT, an event-based asynchronous tracker that represents UAVs as motion-consistent blobs, thereby improving fragmented target handling while suppressing noise and duplicate tracks. We also contribute ES-UAV, a high-definition event-level benchmark for UAV perception. Experimental results confirm a favorable accuracy–efficiency trade-off, with extremely sparse or near-stationary UAVs remaining as open challenges.

\bibliography{aaai2027}

\end{document}